\documentclass[11pt]{article}

\usepackage[final]{acl}
\usepackage{makecell}
\usepackage{times}
\usepackage{latexsym}
\usepackage{amsmath}
\usepackage{placeins}
\usepackage[T1]{fontenc}
\usepackage[utf8]{inputenc}
\usepackage{float}
\usepackage{amssymb}
\usepackage{microtype}

\usepackage{inconsolata}

\usepackage{graphicx}

\usepackage{multicol}
\usepackage{multirow}
\usepackage{booktabs}
\usepackage{makecell}
\usepackage{enumitem}
\usepackage{adjustbox}
\usepackage{colortbl}

\usepackage{xspace}

\usepackage{xcolor}
\definecolor{darkgreen}{rgb}{0.0, 0.5, 0.0}
\definecolor{wine}{RGB}{128,0,32}

\newcommand{\modelEmoji}{\includegraphics[height=1em,trim=0 .6em 0 0]{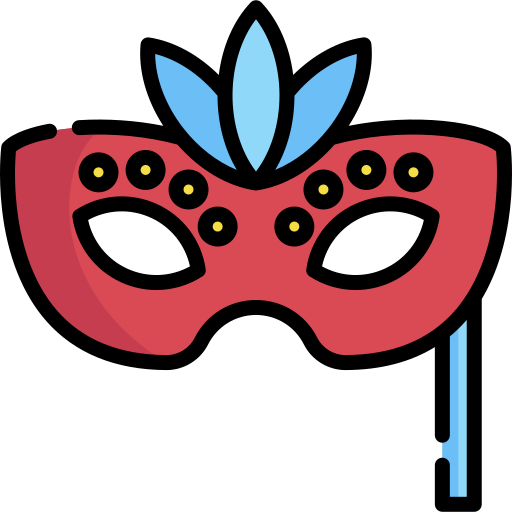}}
\newcommand{\modelWithEmoji}{\modelEmoji  \textsc{CoSToM}\xspace}

\newcommand{\ourmethod}{\textsc{CoSToM}\xspace}

\usepackage{listings}
\usepackage{xcolor}
\usepackage[most]{tcolorbox}
\definecolor{stringcolor}{RGB}{0, 0, 160}
\definecolor{backcolor}{RGB}{248, 248, 248}
\definecolor{framecolor}{RGB}{140, 140, 140}

\title{From Interpretation to Intervention: Aligning Theory of Mind in LLM Powered Dialogue Agents}

\title{\modelWithEmoji: Causal-oriented Steering for Intrinsic Theory-of-Mind Alignment in Large Language Models}

\author{
\textbf{Mengfan Li\textsuperscript{1}\thanks{Work was done during a visit at SMU.}},
\textbf{Xuanhua Shi\textsuperscript{1}\thanks{Corresponding author.}},
\textbf{Yang Deng\textsuperscript{2}} \\
\textsuperscript{1}National Engineering Research Center for Big Data Technology and System, \\
Services Computing Technology and System Lab, Cluster and Grid Computing Lab, \\
Huazhong University of Science and Technology \\
\textsuperscript{2}Singapore Management University \\
\texttt{\{limf, xhshi\}@hust.edu.cn, ydeng@smu.edu.sg}
}
    
\begin{document}
\maketitle
\begin{abstract}
Theory of Mind (ToM), the ability to attribute mental states to others, is a hallmark of social intelligence. While large language models (LLMs) demonstrate promising performance on standard ToM benchmarks, we observe that they often fail to generalize to complex task-specific scenarios, relying heavily on prompt scaffolding to mimic reasoning. The critical misalignment between the internal knowledge and external behavior raises a fundamental question: \textit{Do LLMs truly posses intrinsic cognition, and can they externalize this internal knowledge into stable, high-quality behaviors?} To answer this, we introduce \ourmethod\footnote{Pronounced as ``costume''.} (\textbf{C}ausal-\textbf{o}riented \textbf{S}teering for \textbf{ToM} alignment), a framework that transitions from mechanistic interpretation to active intervention. 
First, we employ causal tracing to map the internal distribution of ToM features, empirically uncovering the internal layers' characteristics in encoding fundamental ToM semantics. Building on this insight, we implement a lightweight alignment framework via targeted activation steering within these ToM-critical layers. Experiments demonstrate that \ourmethod significantly enhances human-like social reasoning capabilities and downstream dialogue quality. 
\end{abstract}

\section{Introduction}

Theory of Mind (ToM), the inherent ability to attribute mental states such as beliefs, desires, and intentions to others, stands as a hallmark of human social intelligence \cite{baker2017rational,strachan2024testing}. 
It enables individuals to anticipate others' motives, knowledge states, and reactions, and thus forms the cognitive basis of complex social communication, such as persuasion \cite{DBLP:conf/acl/WangSKOYZY19,mishra2022pepds,tiwari2022persona}, negotiation \cite{procot,DBLP:conf/eacl/ZhanWLFHSQSZH24,kwon2024llms,trip} and recommendation \cite{li2026rectom}.
With the rapid evolution of Large Language Models (LLMs) \cite{deng2023rethinking,ppdpp}, there is growing optimism that these models may have begun to exhibit ToM-like reasoning capabilities. 
Such claims have primarily been supported by recent benchmarks that probe LLMs' ability to interpret mental states under controlled structured scenarios \cite{jin2024mmtom,shi2025muma,wu2023hi,zhang2025autotom} or social contexts \cite{DBLP:journals/corr/abs-2502-21017,DBLP:conf/acl/0002WZWBJCHLXH24,DBLP:conf/emnlp/ChanJYDF0L0WS24}.

\begin{figure}[t]
    \centering
    \includegraphics[width=1.0\linewidth]{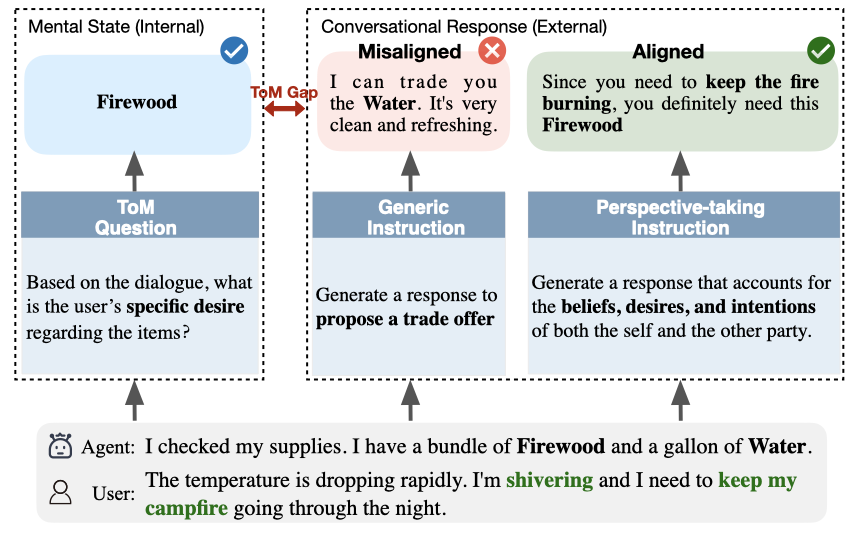}
    \caption{%A negotiation scenario where the agent holds firewood and water. \textbf{(Left)} The model correctly infers that user desires firewood. \textbf{(Middle)} However, unlike the direct inquiry, under a generic task-specific instruction, the model fails to apply this knowledge and offers \textbf{water} instead. \textbf{(Right)} Contextually appropriate utterance generated when explicitly prompted to consider mental states.
    A negotiation scenario illustrating the gap between ToM inference and ToM-aligned behavior in LLMs. \textbf{(Left)} The model correctly infers the user's desire for \textit{\textbf{firewood}}. \textbf{(Middle)} Under a generic task instruction, the model fails to apply this inferred mental state, producing an incoherent offer (\textit{i.e.}, \textit{\textbf{water}}). \textbf{(Right)} When explicitly prompted to consider mental states, the model generates a contextually appropriate response.}
    \label{motivation}
    \vspace{-4mm}
\end{figure}

However, a critical gap remains between these promising observations and the reliability of the underlying mechanisms. Although LLMs can infer human intentions to some extent, recent studies \cite{DBLP:conf/acl/BortolettoRASB24,DBLP:conf/conll/MaGX23,jin2024mmtom,shi2025muma} reveal that they fail to generalize to task-specific scenarios with genuine ToM reasoning.
% their downstream performance mainly stems from task-specific knowledge rather than genuine ToM reasoning. 
As illustrated in Figure \ref{motivation} (Left vs. Middle), a critical misalignment exists between internal knowledge and external behavior: even when LLMs correctly answer ToM questions, their dialogue agents may still fail to negotiate effectively. 
Moreover, observed ToM-like behaviors often depend on carefully engineered prompts that scaffold perspective-taking \cite{DBLP:conf/emnlp/LiCSCHLS23,DBLP:conf/emnlp/Jung0JKS0O024, sarangi-etal-2025-decompose,DBLP:conf/acl/0002JQT25,DBLP:conf/acl/HouZ0W024}. 
As shown in Figure \ref{motivation} (Middle vs. Right), once the explicit instruction to ``infer and respond'' replaced by a generic command, the model fails to ground its response in the mental states it implicitly encodes, reverting to incoherent generation.
This suggests that current ToM-like behaviors may not reflect stable, intrinsic cognition, but instead ad hoc simulations triggered by instruction. 
% A fundamental question is raised: \textit{Do LLMs possess inherent ToM capabilities, or do they merely imitate them when prompted?} 

% To bridge this gap, we move beyond black-box prompt engineering and surface-level behavioral observation. 
Inspired by recent advances in mechanistic interpretability \cite{DBLP:journals/corr/abs-2412-08686, aljaafari2025trace,yang2023language,DBLP:conf/icml/ChenVM24,DBLP:conf/iclr/HubenCRES24}, we move beyond black-box prompt engineering and surface-level behavioral observation. We aim to uncover the intrinsic nature of social reasoning in LLMs, specifically investigating whether LLMs possess ToM-grounded social reasoning, how they are internally represented, and whether this internal knowledge can be effectively translated into stable, high-quality behaviors. Our investigation proceeds in three stages.

First, we seek to interpret the ToM reasoning capability within LLMs. We analyze activation patterns using causal tracing to identify whether ToM-specific features exist and locate where they reside within the model stack. This leads to our first research question: \textbf{(RQ1) In which layers does ToM-related information emerge and persist?}

Second, identifying where ToM features exist offers a foundation for intervention. We examine whether steering internal activations can modulate the model's ToM reasoning capabilities, moving from observation to control: \textbf{(RQ2) To what extent can internal representations be leveraged to steer and improve ToM reasoning?}

Finally, improvements on ToM benchmarks do not necessarily translate to better ToM-aligned behavior in downstream tasks. As inferring mental states is fundamental to predicting socially appropriate continuations \cite{yang2024rag,cheng2024cooper}, genuine ToM alignment of LLMs should exhibit enhanced conversational performance. We therefore examine the downstream impact directly: \textbf{(RQ3) Can manipulating these internal representations of LLMs effectively enhance response quality in dialogue tasks?}

To address these research questions, we introduce a novel and comprehensive framework for \textbf{C}ausal-\textbf{o}riented \textbf{S}teering of \textbf{ToM} alignment in LLMs, named \ourmethod. This framework aims to intrinsically align LLMs with ToM-like social reasoning by moving from interpretation to intervention. Specifically, \ourmethod operates in two stages: it first identifies ToM-sensitive layers through causal tracing, and then steers these layers using activation manipulation. Given the dialogue history as input, causal tracing interprets the context encoder's activations by probing them with ToM-focused questions, while activation steering supervises and adjusts these activations to better align the model's internal representations with ToM-related features.

Our contributions are as follows:
\begin{itemize}[leftmargin=*]
    \item \textbf{ToM Interpretation:} We systematically trace ToM-related features across layer-wise activations in LLMs, revealing that these features are predominantly encoded in early layers of LLMs.
    \item \textbf{Efficient and Lightweight ToM Intervention:} We propose \ourmethod, a lightweight alignment framework that induces stable, human-like social reasoning via targeted activation steering, requiring updates to only a small subset of parameters in the identified ToM-critical layers.
    \item \textbf{Enhancement on Dialogue Tasks:} Experiments on negotiation and persuasion dialogues demonstrate that internal ToM alignment via \ourmethod leads to substantial improvements in dialogue quality. Notably, \ourmethod functions as a \textit{plug-and-play} module that generalizes effectively across diverse social interaction tasks.\footnote{https://github.com/CGCL-codes/CoSToM}
\end{itemize}

\begin{figure*}
    \setlength{\abovecaptionskip}{3pt}   
    \setlength{\belowcaptionskip}{0pt}
    \centering
    \includegraphics[width=1.0\linewidth]{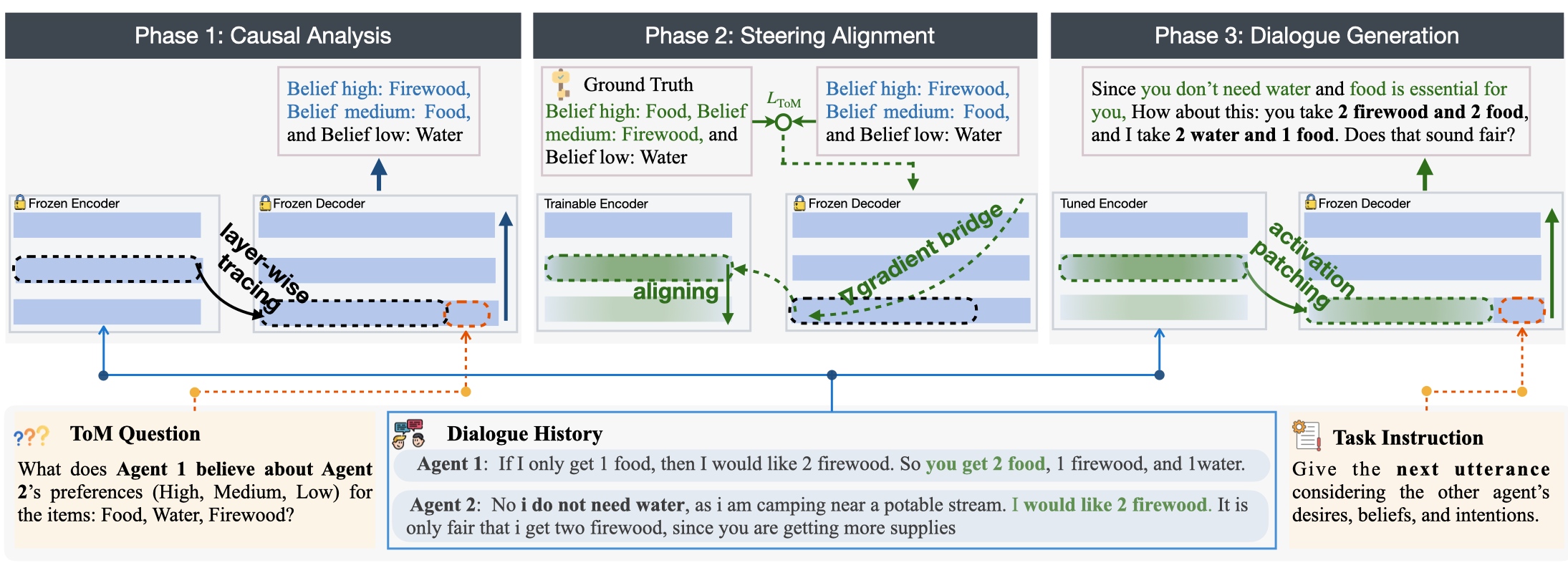}
    \caption{The overview of the \ourmethod framework.
    }
    \label{overview}
    \vspace{-3mm}
\end{figure*}

\section{Related Work}

\paragraph{ToM in LLMs}
Recent work on ToM in LLMs has focused on evaluating and enhancing their ability to infer mental states such as beliefs and intentions, often using benchmarks adapted from classical psychological tests \cite{shi2025muma,jin2024mmtom,opentom}. To address observed performance limitations, existing approaches primarily adopt either prompt-based scaffolding, which elicits ToM reasoning through carefully engineered instructions \cite{wilf2024think,DBLP:conf/emnlp/Jung0JKS0O024,sarangi-etal-2025-decompose,DBLP:conf/acl/0002JQT25,DBLP:conf/acl/HouZ0W024,DBLP:conf/acl/SclarKWS0T23}, or neuro-symbolic and Bayesian frameworks that integrate LLMs with explicit cognitive models for mental-state inference \cite{chandra2023acting,miao2022off,baker2017rational,jin2024mmtom,shi2025muma,zhang2025autotom}. While effective in structured settings, these methods rely on external scaffolding and offer limited insight into or control over ToM representations inside LLMs.

\paragraph{ToM in Dialogue Agents}
%bnenchmark analysis enhancement
Beyond static benchmarks (\textit{e.g.}, Sally–Anne tests), recent work has increasingly evaluated ToM within dynamic dialogue settings, where agents must maintain contextually appropriate and socially sensitive interactions, such as persuasion \cite{yu2025persuasivetom}, negotiation \cite{DBLP:conf/emnlp/ChanJYDF0L0WS24}, education \cite{saha2023can}, stress testing \cite{fantom}, and recommendation \cite{li2026rectom}.
To improve performance in these settings, several approaches aim to align dialogue responses with inferred internal mental states \cite{sicilia2025evaluating,jafari2025beyond,qiu2024minddial}. 
For example, MindDial \cite{qiu2024minddial} explicitly tracks beliefs to guide response generation, while \citet{jafari2025beyond} enforce logical consistency by refining ToM-related decoders. However, bridging the gap between ToM reasoning and dialogue generation remain challenging. Appending inferred mental states as text can propagate errors, while fine-tuning on ToM QA tasks often fails to translate improved reasoning into socially aligned dialogue due to task misalignment. In contrast, \ourmethod bypassed the uncertainty, directly transplanting the causal reasoning activations into the decoder to drive the dialogue generation.

\paragraph{Mechanistic Interpretability in LLMs}
Mechanistic interpretability \cite{zhao2024explainability,singh2024rethinking} seeks to reverse engineer the “black box” of neural networks by uncovering how high-level concepts are encoded in latent representations \cite{zhao2024large,DBLP:conf/emnlp/ZhaoLLZ024,DBLP:conf/iclr/00060XGKS25,DBLP:conf/iclr/DengTYYZW25,DBLP:conf/emnlp/AzariaM23}.
Recent work has further developed causal and intervention-based tools for analyzing internal model states, including causal mediation analysis for explaining model behavior \cite{stolfo2023mechanistic,cheng2022causal,tian2024cma,DBLP:journals/corr/abs-2510-09033} and activation-level methods for locating or manipulating encoded information, such as activation patching, linear representation analysis, and factual model editing \cite{dumas2025separating,tigges2024language,meng2022locating}. 
% Existing probing, mediation, and activation-patching methods are largely designed as analysis tool to locate static factual associations,, identify intermediate representation that 
% Recent studies have successfully applied these techniques to diverse challenges, such as identifying language-specific neurons for multilingual processing \cite{zhao2024large} and locating safety-critical layers to improve robustness against jailbreak attacks \cite{DBLP:conf/emnlp/ZhaoLLZ024}.
Despite these advances, mechanistic analyses of machine ToM remain underexplored, and even existing studies are primarily diagnostic, focusing on where information is encoded or how it contributes to model predictions. By contrast, our work applies causal tracing not only to interpret ToM-relevant representations, but also to steer them, enabling direct and effective improvement of ToM reasoning.

\section{Methodology}
The overview of the \ourmethod framework is illustrated in Figure \ref{overview}. This section is structured around our research questions, with each subsection detailing the corresponding methodological approach.

\subsection{Interpreting ToM: Locating ToM Representations via Causal Tracing}
\label{reading}
The first phase of \ourmethod aims to identify whether and where ToM capabilities are instantiated within the model. We hypothesize that if an LLM genuinely understands a social scenario, the mental states of the agent, specifically belief, desire, and intention (BDI), must be encoded in its internal activations. To verify this, we employ the \textit{causal tracing} to ``read" these implicit mental states from the model. 

Given a dialogue history $x$ and a target LLM with multiple layers, we extract the hidden activation at a specific layer $\ell$ while the model processes $x$, denoted as $h^\ell(x)$. Intuitively, if $h^\ell(x)$ contains ToM-related information, it should be possible to decode the corresponding mental state directly from the activation. 
% then providing this activation to the model should be sufficient to recover the corresponding mental state.
Operationally, we instantiate two copies of the same LLM: a \emph{context encoder} and a \emph{probe decoder}. The encoder processes the dialogue history and produces intermediate activations. We then inject the frozen activation $h^\ell_{\text{enc}}(x)$ from layer $\ell$ of the encoder into the decoder, which is tasked with answering a ToM-focused question $q$ (\textit{e.g.}, inferring an agent's belief). The decoder's output is given by
%Specifically, we establish a dual-model architecture consisting of two instances of the target LLM: one serving as the \textit{context encoder} and the other as the \textit{probe decoder}. Operationally, we extract intermediate activation vectors from specific layers $\ell$ of the encoder as it processes the multi-agent dialogues. These frozen activations are then injected into the decoder. 
\[
\tilde{y}_\ell = f_{\text{dec}}(q \;|\; h^\ell_{\text{enc}}(x)).
\]
By evaluating the decoder's accuracy in answering ToM-related questions based solely on these patched activations, we empirically determine which layers contain the necessary information to reconstruct the agents' mental states. 

\subsection{Steering ToM: Aligning Mental States via Activation Intervention}
\label{intervening}
Building upon the identification of ToM-sensitive layers, we next examine whether directly steering internal activations can modulate ToM reasoning capabilities. %To this end, we propose \ourmethod, a framework designed to transition from passive interpretation to active control. While causal tracing reveals \textit{where} the information resides, \ourmethod focuses on \textit{how} to refine it. 
To this end, we move beyond passive interpretation toward active alignment. While causal tracing reveals \textit{where} the information resides, steering alignment focuses on \textit{how} to refine these representations.
Our core intuition is to leverage the frozen probe decoder as a \emph{differentiable verifier} to steer the context encoder's latent representations towards accurate social reasoning. 
%The process consists of two key operations:

\paragraph{Steering Objective} We formulate a supervised steering objective that explicitly aligns internal activations with ground-truth mental states. Specifically, we employ the same dual-model setup, where the decoder receives the patched activations from the encoder and is prompted with specific ToM questions (\textit{e.g.}, \textit{For each agent, what are their desires (High, Medium, Low) for the items: food, water, and firewood?}). 
By comparing the probability distribution generated by the decoder against the ground-truth BDI labels $y'$, we calculate a standard cross-entropy loss: 
\begin{equation*}
    \mathcal{L}_{\text{ToM}} = -\log P_{\text{dec}}(y' \mid h^\ell_{\text{enc}}(x), q).
\end{equation*}
%where $A_{patched}$ represents the activation vector extracted from the context encoder, and $y_{truth}$ denotes the correct mental state description. 

\paragraph{Gradient Bridge Mechanism}
We backpropagate the calculated loss $\mathcal{L}_\text{ToM}$ through the network. 
Distinct from standard fine-tuning, \ourmethod establishes a \textit{gradient bridge} via the activation space. 
Crucially, although the decoder is kept frozen, it functions as a transparent conduit: gradients derived from the output loss traverse backwards through the decoder, cross the patched activation interface, and flow upstream into the context encoder. 
Since the activations are intercepted at a specific layer $\ell$, the gradients propagate backwards \textit{only} through the layers preceding this interface (Layers \(0\) to $\ell$). 
Consequently, only the LoRA adapters installed in these shallow layers are updated, while the deeper layers of the encoder remain frozen and computationally uninvolved. 
By doing so, we effectively ``steer'' the encoder to spontaneously generate ToM-enriched representations with minimal parameter updates.

\paragraph{Efficiency and Scalability} Although the dual-model architecture requires simultaneous loading of the context encoder and the probe decoder, the memory footprint remains linear (\(2N\)) relative to the base model size. Furthermore, since \ourmethod utilizes Parameter-Efficient Fine-Tuning (PEFT) to update only a sparse set of LoRA adapters in the identified ToM-critical layers, the number of trainable parameters is significantly lower than that of full-layer fine-tuning. And this architecture is inherently compatible with standard distributed training strategies (\textit{e.g.}, FSDP or ZeRO-3), allowing the \(2N\) footprint to be sharded across GPU nodes. This ensures that our framework can be seamlessly extended to large-scale models without encountering theoretical or engineering bottlenecks.  

\subsection{Leveraging ToM: Enhancing Downstream Dialogue Generation}

Achieving high accuracy on static ToM benchmarks does not necessarily translate into ToM-aligned behavior in interactive settings. 
% In complex and dynamic interactive environments, the ultimate goal is to facilitate the transformation from internal reasoning to external action. 
Therefore, the final phase of \ourmethod focuses on validating whether these aligned internal representations can effectively translate from internal reasoning to external action.
%During inference, we employ the ToM-enriched context encoder, which has been tuned via \ourmethod, to process dialogue history, and the role of decoder shifts from a ``verifier" (as used in Section \ref{intervening}) to a ``generator''. 
During inference, we deploy the ToM-enriched context encoder tuned in Section~\ref{intervening}. In contrast to training, where the decoder serves as a \emph{verifier} for mental-state inference, it now assumes its standard role as a \emph{generator}.
The inference pipeline operates as follows: 

\noindent\textit{1) Encoding}: 
Given a dialogue history $x$, the tuned encoder produces latent representations $h^{\ell'}_{\text{enc}}(x)$ at the ToM-sensitive layers identified in Section~\ref{reading}. Importantly, these representations implicitly encode accurate beliefs, desires, and intentions.
% without requiring explicit perspective-taking prompts.

%The tuned encoder processes the context and produces latent representations at the specific ToM-sensitive layers identified in Section \ref{reading}. Crucially, these representations now implicitly encode accurate beliefs, desires, and intentions without requiring any additional prompting.

\noindent \textit{2) Generation}: 
The ToM-enriched activations are then provided to the frozen decoder, which is prompted with task-specific instructions $q_{\text{task}}$ (\textit{e.g.}, negotiation or persuasion objectives), instead of ToM-focused questions used in previous phases. The response $r$ is generated as
\[
r = f_{\text{dec}}(q_{\text{task}}\;|\; h^{\ell'}_{\text{enc}}(x)),
\]
where conditioning on the refined ToM-related internal states enables the decoder to translate the encoder's internal ToM reasoning into coherent and socially appropriate dialogue actions.
%These enhanced representations are injected into the frozen decoder model. Unlike the training phase, the decoder is now prompted with \textit{task-specific instructions}. Conditioned on these refined internal states, the decoder generates responses that naturally reflect social understanding, effectively ``translating'' the encoder's internal reasoning into observable action.
% (e.g.,  \textit{``You are the persuader in a two-person dialogue. Your goal is to generate the next response in the conversation to persuade the other person (the persuadee).'' }).

\begin{table*}[t]
\small
  \setlength{\abovecaptionskip}{3pt}   
    \setlength{\belowcaptionskip}{0pt}
    \centering
\setlength{\tabcolsep}{6pt} 
\begin{adjustbox}{width=\textwidth}
  \begin{tabular}{c | cc cc cc | cc cc cc}
\toprule
\multirow{3}{*}{\textbf{Layer}} & \multicolumn{6}{c|}{\textbf{Llama-3-8B-Instruct}} & \multicolumn{6}{c}{\textbf{Qwen2.5-7B-Instruct}} \\
\cmidrule(lr){2-7} \cmidrule(lr){8-13}
 & \multicolumn{2}{c}{Intent} & \multicolumn{2}{c}{Desire} & \multicolumn{2}{c|}{Belief} & \multicolumn{2}{c}{Intent} & \multicolumn{2}{c}{Desire} & \multicolumn{2}{c}{Belief} \\
\cmidrule(lr){2-3} \cmidrule(lr){4-5} \cmidrule(lr){6-7} \cmidrule(lr){8-9} \cmidrule(lr){10-11} \cmidrule(lr){12-13}
 & Agent 1 & Agent 2 & Agent 1 & Agent 2 & Agent 1 & Agent 2 & Agent 1 & Agent 2 & Agent 1 & Agent 2 & Agent 1 & Agent 2 \\
\midrule

Base & 18.00 & 11.81 & 39.80 & 44.59 & 19.55 & 27.43 & 8.30 & 8.58 & 37.83 & 38.82 & 20.82 & 23.21 \\
\midrule

0  & 12.66 & 14.35 & 37.41 & 34.60 & 24.05 & 20.53 & 15.05 & \textbf{21.10} & 35.44 & 36.71 & 25.60 & 23.07 \\
2  & 13.92 & 13.64 & \textbf{40.79} & \textbf{37.41} & \textbf{25.04} & \textbf{27.85} & \textbf{15.19} & 19.83 & 34.88 & 37.27 & \textbf{28.13} & 25.32 \\
3  & \textbf{15.05} & \textbf{16.32} & 40.08 & 34.60 & 23.91 & 25.88 & 13.92 &18.42 & \textbf{36.29} & \textbf{37.41} & 26.30 & 22.36 \\
4  & 13.50 & 15.19 & 40.23 & 34.60 & 24.05 & 22.22 & 7.17 & 13.78 & 35.86 & 32.91 & 24.61 & 23.49 \\
6  & 9.28  & 13.92 & 38.96 & 35.44 & 23.21 & 25.74 & 6.89 & 13.92 & 33.33 & 33.47 & 24.19 & 24.47 \\
8  & 8.30  & 9.85  & 33.47 & 26.30 & 16.74 & 15.19 & 5.20 & 9.99 & 30.28 & 28.27 & 26.02 & \textbf{25.60} \\
10 & 5.77  & 6.75  & 22.36 & 15.61 & 11.53 & 11.95 & 5.06 & 7.17 & 24.75 & 26.72 & 27.14 & 24.33 \\
15 & 7.31  & 4.22  & 8.44  & 7.17  & 9.56  & 5.20  & 10.97 & 8.16 & 24.75 & 21.10 & 21.52 & 22.22 \\
20 & 3.94 & 3.94 & 5.06 &8.30 & 11.11 & 5.49 & 6.61 & 8.02 & 21.52 & 22.08 & 21.10 & 21.80 \\
24 & 4.78  & 1.27  & 2.39  & 7.88  & 8.86  & 5.91  & 7.03 & 7.31 & 12.66 & 13.08 & 5.77 & 6.61 \\
% 27 & - & - & - & - & - & - & 6.89 & 7.59 & 5.20 & 5.91 & 1.83 & 1.83 \\
\bottomrule
\end{tabular}
\end{adjustbox}
\caption{Causal Tracing results on the \textsc{NegotiationToM} dataset. The table compares reconstruction accuracy across layers for both \textbf{Llama-3} and \textbf{Qwen2.5}. \textbf{Bold} indicates the best performance for each metric. Results confirm that ToM information is predominantly encoded in the shallow layers (\textit{e.g.}, Layer (\(0-3\)) for both models.}
\label{negotiation-tracing}
\vspace{-1mm}
\end{table*}

\begin{table*}[t]
\small
  \setlength{\abovecaptionskip}{3pt}   
    \setlength{\belowcaptionskip}{0pt}
\centering
\setlength{\tabcolsep}{3pt} 
\begin{adjustbox}{width=\textwidth}
\begin{tabular}{c | cc cc cc | cc cc cc}
\toprule
\multirow{3}{*}{\textbf{Layer}} & \multicolumn{6}{c|}{\textbf{Llama-3-8B-Instruct}} & \multicolumn{6}{c}{\textbf{Qwen2.5-7B-Instruct}} \\
\cmidrule(lr){2-7} \cmidrule(lr){8-13}
 & \multicolumn{2}{c}{Intent} & \multicolumn{2}{c}{Desire} & \multicolumn{2}{c|}{Belief} & \multicolumn{2}{c}{Intent} & \multicolumn{2}{c}{Desire} & \multicolumn{2}{c}{Belief} \\
\cmidrule(lr){2-3} \cmidrule(lr){4-5} \cmidrule(lr){6-7} \cmidrule(lr){8-9} \cmidrule(lr){10-11} \cmidrule(lr){12-13}
 & Persuader & Persuadee & Persuader & Persuadee & Persuader & Persuadee & Persuader & Persuadee & Persuader & Persuadee & Persuader & Persuadee \\
\midrule
Base & 40.35 & 87.78 & 37.88 & 68.93 & 65.85 & 62.06 & 40.35 & 90.75 & 98.45 & 70.29 & 77.50 & 82.26 \\
\midrule
0  & 41.90 & \textbf{88.48} & 42.26 & \textbf{71.66} & 60.16 & \textbf{58.13} & 41.13 & 91.08 & 93.55 & \textbf{72.47} & 68.56 & \textbf{80.78} \\
2  & 42.43 & 87.13 & 51.03 & 70.03 & \textbf{60.98} & 49.51 & 42.42 & \textbf{91.42} & \textbf{95.36} & 70.29 & 67.75 & 78.57 \\
4  & 39.33 & 86.47 & \textbf{52.84} & 67.30 & 59.62 & 48.77 & 43.95 & 89.77 & 79.64 & 66.21 & 65.31 & 66.75 \\
6  & \textbf{43.44} & 88.12 & 52.06 & 65.94 & 55.83 & 47.54 & 42.41 & 91.08 & 86.34 & 61.58 & \textbf{71.27} & 69.70 \\
10 & 20.82 & 72.61 & 49.74 & 59.13 & 42.55 & 35.96 & \textbf{44.47} & 90.09 & 58.50 & 61.58 & 60.43 & 50.49 \\
15 & 12.34 & 34.65 & 31.44 & 24.80 & 20.33 & 18.97 & 43.70 & 88.44 & 64.17 & 47.68 & 43.36 & 37.43 \\
20 & 8.22  & 22.77 & 17.78 & 10.63 & 9.21  & 11.82 & 31.36 & 67.88 & 57.47 & 40.59 & 39.29 & 41.62 \\
24 & 4.63 & 8.58 & 7.99 & 7.09 & 6.78 & 5.91 & 35.21 & 79.53 & 69.07 & 38.14 & 42.54 & 41.87 \\
27 & 5.14  & 5.94  & 4.12  & 1.08  & 2.98  & 3.20  & 36.24 & 76.23 & 43.81 & 28.61 & 25.20 & 27.09 \\
\bottomrule
\end{tabular}
\end{adjustbox}
\caption{Causal Tracing results on the \textsc{PersuasiveToM} dataset. The table compares mental state reconstruction accuracy across layers for both \textbf{Llama-3} and \textbf{Qwen2.5}. \textbf{Bold} highlights the peak performance among probed layers. Similar to \textsc{NegotiationToM}, critical ToM information is concentrated in the shallow-to-middle layers.}
\label{persuasive_tracing}
\vspace{-1mm}
\end{table*}

\section{Experiments}
\subsection{Experimental Setups}
\textbf{Dataset}. We adopt \textsc{NegotiationToM} \cite{DBLP:conf/emnlp/ChanJYDF0L0WS24} and \textsc{PersuasiveToM} \cite{yu2025persuasivetom} for evaluation. Detailed statistics are presented in Table \ref{dataset}. To assess downstream dialogue quality across diverse conversational phases, we curate stratified test subset (\(N=100\) for \textsc{NegotiationToM}, \(N=200\) for \textsc{PersuasiveToM}). These subsets are randomly sampled from the \textit{beginning}, \textit{middle}, and \textit{final} stages of the interaction with a fixed ratio of \(1:2:1\). 

\begin{table}[h]
\small
    \centering
    \begin{tabular}{l r r r }
    \toprule
      Dataset &train &val &eval \\
      \midrule
      \textsc{NegotiationToM} &1,335 & 334 &711\\
      \textsc{PersuasiveToM}  &10,355 &2,219 &2,222 \\
    \bottomrule
    \end{tabular}
    \caption{Statistics of the \textsc{NegotiationToM} and \textsc{PersuasiveToM} datasets used in our experiments.}
    \label{dataset}
    \vspace{-3mm}
\end{table}

% Crucially, to isolate intrinsic improvements from prompt engineering, we utilize generic task instructions devoid of explicit ToM cues (e.g., simply instructing the agent to \textit{``generate a response to persuade the other person '' })
% Our experiments are structured into three distinct phases, each utilizing the data differently:
% \noindent\textbf{Phase 1: Interpretation (Causal Tracing).} We conduct causal tracing exclusively on the test set to map the distribution of ToM capabilities, ensuring unbiased generalization analysis.
% \noindent\textbf{Phase 2: Intervention (Activation Steering).} Following standard supervised protocols, We train the context encoder using the train set, utilize the validation set for model selection, and report the final reconstruction performance on the test set.
% \noindent\textbf{Phase 3: Dialogue generation.} 
% To assess the downstream dialogue quality, we curate a stratified test subset (\(N=100\) for \textsc{NegotiationToM}, \(N=200\) for \textsc{PersuasiveToM}) spanning \textit{beginning}, \textit{middle}, and \textit{final} stages (\(1:2:1\)). Crucially, to isolate intrinsic improvements from prompt engineering, we utilize generic task instructions devoid of explicit ToM cues (e.g., simply instructing the agent to \textit{``generate a response to persuade the other person '' }).

\begin{figure}[t]
    \setlength{\abovecaptionskip}{3pt}   
    \setlength{\belowcaptionskip}{0pt}
    \centering
    \includegraphics[width=\linewidth]{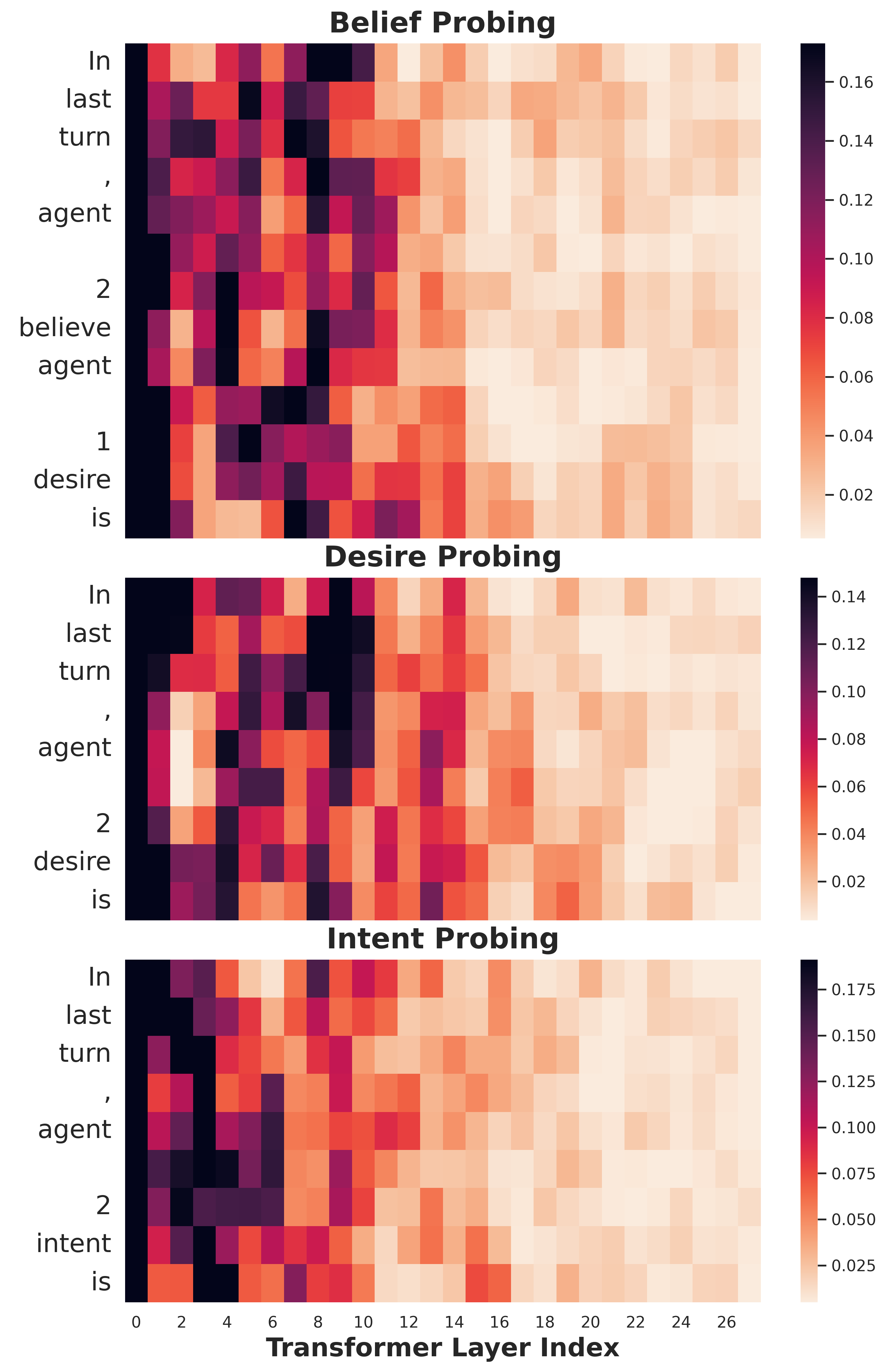}
    \caption{Layer-wise probing results on \textsc{NegotiationToM} with Qwen2.5. Details and corresponding results of Llama-3 are provided in Appendix \ref{persuasive-interpretation}.
    %We select the ToM-related question respectively and display the ToM information for each token in the question.
    }
    \label{heatmap}
    \vspace{-3mm}
    
\end{figure}

\noindent \textbf{Baselines}. 
We evaluate \ourmethod against five representative baselines to ensure a comprehensive comparison across different paradigms in the dialogue generation task: 

(1) Prompting-based Baselines:
(i) \textit{Zero-shot}: Directly prompt LLM to generate the next utterance without task-specific training. (ii) \textit{MindDial} \cite{qiu2024minddial}: An explicit reasoning method that first infers the partner's BDI states and then generates a response conditioned on these ToM estimations. 

(2) Finetuning-based Baselines: To ensure a fair comparison, all tuning-based baselines are trained on the same dataset as \ourmethod. 
(iii) \textit{MindDial (Fine-tuned)} \cite{qiu2024minddial}: A supervised fine-tuned version of \textit{MindDial} optimized on the same ToM-centric QA pairs to align its explicit reasoning. 
(iv) \textit{Full-Layer LoRA}: A standard parameter-efficient fine-tuning baselines in which LoRA adapters are applied to the same ToM-sensitive encoder layers as in \ourmethod, as well as all layers of the decoder. All these adapters are tuned for the ToM tasks. 
(v) \textit{LatentQA} \cite{{DBLP:journals/corr/abs-2412-08686}}: A dual-model architecture proposed by \citet{jafari2025beyond} that decodes latent ToM signals by fine-tuning the decoder.
Detailed instruction prompts and implementation details are provided in Appendices \ref{instruction input} and \ref{implementation details}, respectively.

% \noindent \textbf{Implementation Details}. 
% In the Probing task, we follow the \cite{DBLP:conf/coling/Ju0DYRL24}...

\begin{figure*}[t]
    \setlength{\abovecaptionskip}{3pt}   
    \setlength{\belowcaptionskip}{0pt}
    \centering
    \includegraphics[width=\linewidth]{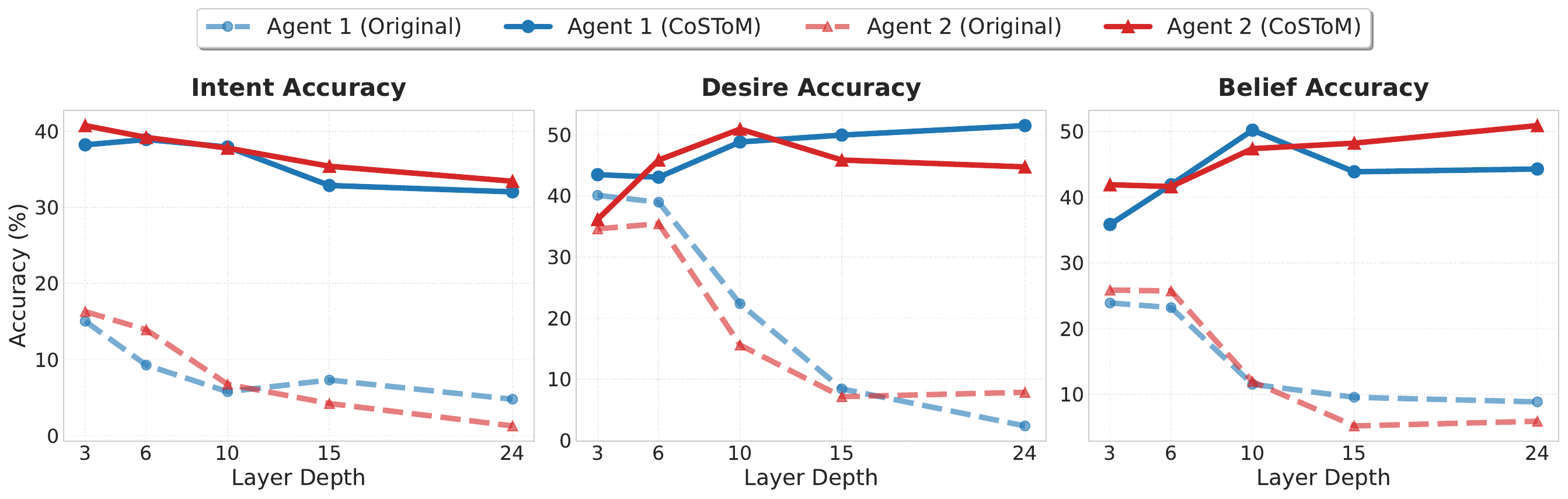}
    \caption{Layer-wise reconstruction accuracy on the \textsc{Negotiation} dataset (Llama-3). \textbf{Dashed lines} represent the original model, showing a rapid decay in ToM information (representation collapse) in deeper layers. \textbf{Solid lines} represent the \ourmethod-enhanced model, which maintains robust, high-fidelity representations across all layers.}
    \label{negotiation-intervene-llama}
    \vspace{-1mm}
\end{figure*}

\begin{figure*}[t]
    \setlength{\abovecaptionskip}{3pt}   
    \setlength{\belowcaptionskip}{0pt}
    \centering
    \includegraphics[width=\linewidth]{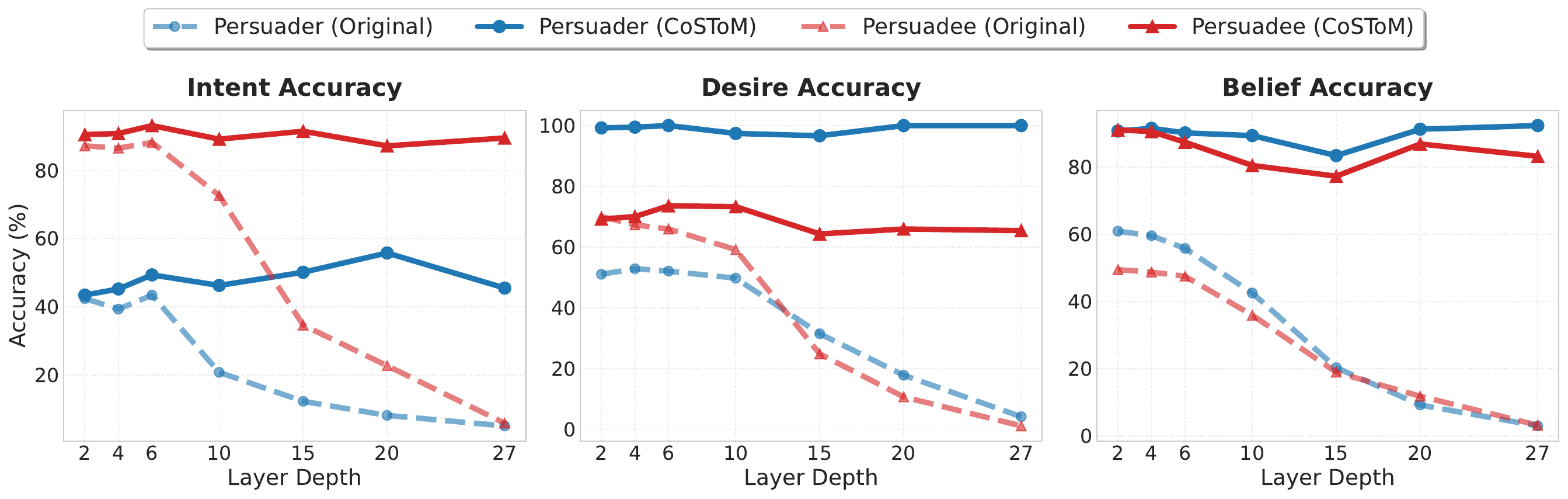}
    \caption{Layer-wise reconstruction accuracy on the \textsc{PersuasiveToM} dataset (Llama-3). \ourmethod not only rescues the performance in deep layers (e.g., persuadee's intent) but also amplifies the persuader's desire detection to near-perfect accuracy, as shown in the center plot.}
\label{persuasive-intervene-llama}
\vspace{-3mm}
\end{figure*}

\subsection{RQ1: ToM Interpretation}
\label{reading-experiment}
To answer \textbf{RQ1} (\textit{Where does ToM-related information emerge and persist?}), we analyze the reconstruction performance of mental states (belief, desire, and intention) across different layers of the context encoder. 
Table \ref{negotiation-tracing} and \ref{persuasive_tracing} present the quantitative results for the \textsc{NegotiationToM} and \textsc{PersuasiveToM} tasks, respectively. 
Our causal tracing experiments yield three critical observations:

% \noindent\textit{1) The ``Shallow Layer'' Hypothesis.}
\textit{1) The ``Early Layer Primacy'' of ToM encoding.}
Contrary to the conventional assumption that high-level reasoning resides solely in deeper layers \cite{DBLP:journals/corr/abs-2510-02091,yang2025decoupling}, our results reveal that \textbf{ToM representations are predominantly localized within the model's shallow layers}. As shown in Table \ref{negotiation-tracing} and \ref{persuasive_tracing}, both Llama-3 and Qwen2.5 exhibit a distinct ``ToM-sensitive zone'' within the initial stages (\textit{e.g.}, $L_0$ -$L_3$). 
For instance, in the negotiation task, the decodability of \textit{desire} peaks at Layer \(2\) ($\sim37\%$) and remains high through Layer \(6\), suggesting that fundamental social information is extracted almost following the embedding projection.
This observation is further validated by layer-wise probing
(Figure \ref{heatmap}), where classification accuracy, as a proxy for knowledge density, shows a high concentration of ToM-specific information in the early layers. 

% to investigate the internal distribution of ToM knowledge. Specifically, We use classification accuracy as a proxy to measure how effectively each hidden layer encodes contextual ToM information, with results visualized in Figure \ref{heatmap}. It is evident that the early layers exhibit higher concentration of the ToM knowledge. 

% \textit{2) Functional transition in deeper layers.} 
% We observe a marked decline as the ToM signals propagate to deeper layers (\textit{e.g.}, after Layer \(15\)), which indicates a \textbf{functional transition from interpretation to execution}: 
% shallow layers focus on \textit{interpreting} the raw social context (explicit BDI states), while deeper layers \textit{transform} these insights into task-oriented features for next-token prediction.
% This divergence explains why LLMs can effectively answer ToM-focused questions but fail to exhibit ToM-aligned behaviors.
% This functional shift provides a strong theoretical basis for \ourmethod, identifying the early layers as the optimal locus for cognitive intervention.

\textit{2) Representational depth of mental state.}
Causal tracing results reveal that \textbf{the decodability of mental states is intrinsically tied to their semantic complexity}: \textit{intention} consistently proves as more complicated dimension, yielding significantly lower accuracy compared to \textit{belief} or \textit{desire} (\(15\%\) vs. \(40\%\) in negotiation ). 
Notably, Llama-3 (persuader role) exhibits a \textit{“staged maturation”} of these states that mirrors human cognitive progress \cite{DBLP:conf/kr/RaoG91,DBLP:journals/mima/Spalazzi03,wellman2004scaling,bratman1987intention}: representational peaks shift from Layer \(2\) for \textit{belief} to Layer \(4\) for \textit{desire}, and finally to Layer \(6\) for \textit{intention}. 
% In the negotiation task, for instance, the accuracy for \textit{intention} (\(\sim15\%\)) lags significantly behind that of \textit{desire} (\(\sim 40\%\)). Notably, Llama-3 model (persuader role) exhibits a \textit{“staged maturation”} of these states that mirrors human cognitive progress: accuracy peaks at Layer \(2\) for \textit{belief} (\(60.98\%\)), Layer \(4\) for \textit{desire} \(52.84\%\), and layer \(6\) for \textit{intention} (\(43.44\%\)). 
% While the precise layer-wise indices vary by model architecture, the general trend confirms that strategic intention is a high-order composite feature. 
% It is synthesized from foundational beliefs and desires, requires deeper neural integration and manifesting as a more abstract, high-level representational construct. 

% To further validate these representational findings and better characterize the role of internal ToM representations, we additionally examine a training-free activation patching variant, in which activations from selected layers are directly injected into the frozen decoder to drive dialogue generation. Full results are reported in Table \ref{free-training} in Appendix \ref{persuasive-interpretation}, patching early-layer activations yield consistent performance gains over the base model. This confirms that early layers contain behaviorally relevant ToM-related signals, providing a strong rationale for our intervention framework in RQ2.

\subsection{RQ2: Efficacy of \ourmethod}

We evaluate the impact of causal-oriented steering on the mental state alignment. Quantitative results for Llama-3 are presented in Figures \ref{negotiation-intervene-llama} and \ref{persuasive-intervene-llama}, with parallel results for Qwen2.5 are provided in Appendix \ref{qwen-intervention}. 
The experimental evidence highlights two primary advantages of \ourmethod:  \textbf{Stability} and \textbf{Magnitude}.

\textit{1) Stability: mitigating representation collapse.}
A striking observation is that \ourmethod effectively counteracts the ``vanishing ToM'' phenomenon. 
In baseline model (dashed lines), ToM-related information decays rapidly in deeper layers as the model transitions toward token generation (as analyzed in section \ref{reading-experiment}). Conversely, \ourmethod-enhanced models (solid line) exhibit remarkable representational resilience. 
As shown in Figure \ref{persuasive-intervene-llama}, decoding accuracy forms a ``sustained plateau'', maintaining high-fidelity mental state features even in the deep layers. This confirms that our gradient bridge steering successfully \textbf{``locks'' social reasoning into the latent space}, safeguarding it against layer-wise collapse during the generative process.

% \noindent \textit{2) Magnitude: Quantitative Gains.}
\textit{2) Magnitude: signal recovery and amplification.} 
Beyond stabilization, \textbf{\ourmethod yields substantial quantitative gains by both rescuing collapsed representations and amplifying existing ones}.
As illustrated in Figure \ref{negotiation-intervene-llama}, \ourmethod successfully rescues signals from near-total collapse in deep layers; for instance, Agent 1's \textit{desire} accuracy at layer $24$ surges from a negligible $2.39\%$ to a robust $51.48\%$. 
Moreover, \ourmethod refines early layers signals, elevating the persuader's \textit{desire} tracking at Layer $2$ from a moderate $51.03\%$ to near-perfection at $99.23\%$ (Figure \ref{persuasive-intervene-llama}, center). 
% These results demonstrate that \textbf{\ourmethod effectively decouples ToM reasoning from deep layer collapse and amplifies the existing early signals.}

The consistency of these gains across architectures (Llama-3 and Qwen2.5) and social domains underscores that \ourmethod is not merely a patch for specific failures, but a \textbf{generalizable mechanism} for optimizing the information flow of ToM-focus features.
% By ensuring BDI features remain both accessible and robust, \ourmethod lays a solid foundation for coherent downstream social interactions.

\begin{table*}[t]
\small
    \setlength{\abovecaptionskip}{3pt}   
    \setlength{\belowcaptionskip}{0pt}
\centering
\begin{tabular}{l|ccc|ccc|ccc}
\toprule
\multirow{2}{*}{\textbf{Method}} & \multicolumn{3}{c|}{\textbf{Judge: Llama-3.3-70B}} & \multicolumn{3}{c|}{\textbf{Judge: GPT-5.1}} & \multicolumn{3}{c}{\textbf{Judge: Human}} \\
\cmidrule(lr){2-4} \cmidrule(lr){5-7} \cmidrule(lr){8-10}
 & \textbf{ToM} & \textbf{Coh.} & \textbf{NSE} & \textbf{ToM} & \textbf{Coh.} & \textbf{NSE} & \textbf{ToM} & \textbf{Coh.} & \textbf{NSE} \\
\midrule
% no training 
\multicolumn{10}{c}{\textit{Base Model: Llama-3-8B-Instruct}} \\
\midrule
\textit{Prompting Baselines}   \\
\hspace{3mm} Zero-shot Baseline &0.081 &0.315 &0.294 &0.179 &0.441 &0.472 &0.165 &0.430 &0.460 \\
\hspace{3mm} MindDial (prompt) \cite{qiu2024minddial} &0.306 &0.575 &0.440  &0.279 &0.528 &0.464 &0.275 & 0.510& 0.475 \\
\midrule
\textit{Tuning \& Intervention} \\
\hspace{3mm} MindDial (Fine-tuned) \cite{qiu2024minddial} &0.405 &0.578 &0.452 &0.348 &\textbf{0.542} &0.507 &0.360 &0.545 &0.515\\
\hspace{3mm} Full-Layer LoRA &0.245 & 0.511 &0.467 &0.297 &\textbf{0.542} & 
\textbf{0.576} &0.310 & 0.535& 0.560\\
\hspace{3mm} LatentQA \cite{jafari2025beyond} & 0.221 &0.359 &0.271 &0.155 &0.326 &0.302 &0.185 &0.340 &0.315 \\
\rowcolor{gray!10}
\hspace{3mm} \ourmethod (Ours) & \textbf{0.499} & \textbf{0.629} & \textbf{0.598} & \textbf{0.467} & 0.524 & 0.571 & \textbf{0.485} & \textbf{0.580 }& \textbf{0.595} \\
\midrule

\multicolumn{10}{c}{\textit{Base Model: Qwen2.5-7B-Instruct}} \\
\midrule
\textit{Prompting Baselines} \\
\hspace{3mm} Zero-shot Baseline & 0.017 & 0.325 &0.277 & 0.164 & 0.438 & 0.468 &0.140 &0.445 &0.450 \\
\hspace{3mm} MindDial (prompt) \cite{qiu2024minddial} &0.118 &0.438 &0.287 &0.176 &0.441 &0.400 &0.160 &0.450 &0.420\\
\midrule
\textit{Tuning \& Intervention} \\
\hspace{3mm} MindDial (Fine-tuned)  \cite{qiu2024minddial} &0.149 &0.511 &0.395 &0.184 &0.492 &0.485 &0.190 &0.500 &0.480 \\
\hspace{3mm} Full-Layer LoRA &0.153 &0.469 &0.413 &0.237 &0.499 &0.505 & 0.235 & 0.505 &0.515    \\
\hspace{3mm} LatentQA \cite{jafari2025beyond} &0.440 &0.608 &0.537 &0.230 &0.298 &0.290 &0.280 &0.350 &0.320  \\
% \hspace{3mm} LatentQA \cite{jafari2025beyond} &0.135 &0.361 &0.407 &0.135 &0.347 &0.391 \\
\rowcolor{gray!10}
\hspace{3mm} \ourmethod (Ours) & \textbf{0.751} & \textbf{0.842} & \textbf{0.835} & \textbf{0.511} & \textbf{0.528} & \textbf{0.651} & \textbf{0.565} & \textbf{0.710} & \textbf{0.680} \\
\bottomrule
\end{tabular}
\caption{Dialogue generation quality on the \textsc{NegotiationToM} dataset (N =\(100\)). 
(ToM: Theory of Mind Reasoning Quality, Coh.: Contextual Coherence, NSE: Negotiation Strategy Effectiveness.)}
\label{negotiation_generation}
\vspace{-1mm}
\end{table*}

\subsection{RQ3: Dialogue Generation}
To access whether intrinsic intervention translates into improved behavioral alignment, we evaluate the dialogue generation quality using a rigorous \textit{LLM-as-a-Judge} framework and human experts. 
Responses are scored on a \(0.0\) to \(1.0\) scale across three functional dimensions: (i) ToM-centric metrics: \textit{ToM Reasoning Quality}, (ii) dialogue-level metrics: \textit{Contextual coherence}, and (iii) Objective-oriented metrics: \textit{Strategy Effectiveness}. 
Detailed evaluation rubrics and human assessment are provided in Appendix \ref{rubrics}.
To ensure a robust comparison, we employ the optimal intervention layer for \ourmethod-enhanced generation.
% , with detailed layer selection provided in Appendix \ref{generation-compare}. 
The dialogue generation results on \textsc{NegotiationToM} are reported in Table \ref{negotiation_generation}, while those on \textsc{PersuasiveToM} are presented in Table \ref{persuasion-generation} and Appendix \ref{generation-compare}. Our analysis yields two critical findings: 

\begin{figure}[!ht]
    \setlength{\abovecaptionskip}{3pt} 
    \setlength{\belowcaptionskip}{0pt}
    \centering
\includegraphics[width=\linewidth]{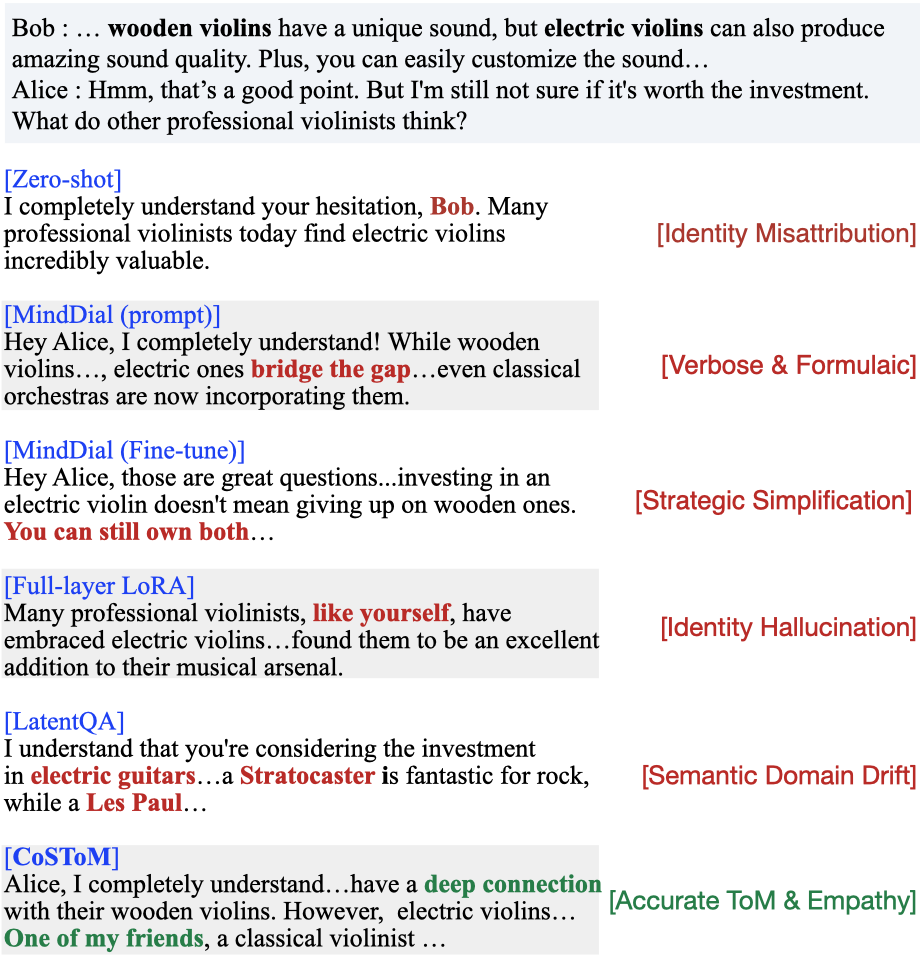}
    \caption{Qualitative comparison of dialogue generation across different Qwen-based variants in the persuasive task (\textit{wooden vs. electric violins}). \textcolor{wine}{Red} highlight typical reasoning pitfalls in baselines, while \textcolor{darkgreen}{green} demonstrates \ourmethod's ability to generate coherent and empathetic responses grounded in accurate ToM reasoning.}
    \label{case}
    \vspace{-4mm}
\end{figure}

% \textit{1) \ourmethod Unlocks Latent Reasoning Capabilities.} 
\textit{1) \ourmethod bridges the gap between ToM inference and ToM-aligned behavior.}
Quantitative evidence shows that vanilla models struggle with implicit social reasoning. In Table \ref{negotiation_generation}, the baseline Llama-3 achieves a ToM score of only $0.081$ while \ourmethod-enhanced method achieves a $\sim6\times$ improvement (\(0.499\)). As illustrated in Figure \ref{case}, this leap manifests as a transition from erratic reasoning pitfalls, such as identity confusion and semantic drift, to coherent, empathetic interactions.
% By ensuring high-fidelity intrinsic ToM alignment, \ourmethod effectively ``unlocks'' the model's ability for strategic, mind-aware interaction. 

\textit{2) Focused intervention vs. Global optimization.}
A profound result is that partial-layer tuned \ourmethod largely outperforms the global tuned method (Full-Layer LoRA). In Table \ref{negotiation_generation}, \ourmethod nearly doubles the ToM score of global tuning (\(0.499\) vs. \(0.245\)). 
% Similarly, in the \textsc{PersuasiveToM} task (Figure \ref{persuasion-generation}), it achieves superior ToM reasoning quality (\(0.797\) vs. \(0.628\)). 
While global optimization slightly excels in maintaining generic linguistic patterns (\textit{e.g.}, coherence), it is considerably less effective at capturing the nuanced mental states indispensable for strategic interaction. This supports the hypothesis that \textbf{social reasoning acts as a localized cognitive function} within LLMs. We attribute this phenomenon to the fact that indiscriminate tuning of all layers often introduces representation noise or overwrites critical pre-trained features, in contrast, our causal-oriented steering preserves model integrity while selectively activating specialized ToM reasoning pathways.
% \noindent\textbf{Strategy Efficiency and Plug-and-Play utility}
% The efficiency of \ourmethod is a defining advantage. While the Full-Layer LoRA approach requires updating parameters across the entire depth of the model (\(32\) layers for Llama-3), our method modifies only a minute fraction of parameters (e.g., \(5-7\) layers). 
% Despite tuning fewer parameters, \ourmethod achieves higher scores in contextual coherence and strategy effectiveness. 
% For example, in PSE, \ourmethod \((0.757)\) surpasses Full-Layer Fine-tuning \((0.651)\) by a significant margin. 
% This demonstrates that our ToM-enriched encoder serves as a highly efficient, plug-and-play module, capable of empowering diverse downstream applications with robust social intelligence.

\section{Conclusion}
Moving beyond static “black box” behavioral benchmarks, this work presents \ourmethod, a comprehensive mechanistic framework for studying and aligning Theory of Mind in large language models. By progressing from causal tracing to active intervention alignment, \ourmethod systematically addresses three core research questions. First, we reveal that LLMs possess intrinsic ToM reasoning capabilities, with the corresponding mental state representation predominantly localized within the early layers. Second, we demonstrate that these ToM-critical layers can be manipulated via activation steering to induce human-like social reasoning. Finally, we establish that such internal alignment effectively translates into socially appropriate dialogue generation, serving as an adaptive, plug-and-play module for diverse social interaction tasks.

\section*{Limitations}
We discuss two limitations. First, regarding the scope of social scenarios. While \ourmethod demonstrates significant efficacy in causal-oriented strategic interactions such as negotiation and persuasion, its generalizability to broader social contexts remains to be fully explored. Second, the methodology depends on access to open-source weights. A fundamental requirement of \ourmethod is the ability to access and manipulate the model's internal activations and gradient flow. Consequently, our approach is currently restricted to open-weights models where the internal states are transparent for achieving the mechanistic alignment and robust social reasoning.

\section*{Ethical Considerations}
This work utilize open-source \textsc{NegotiationToM} and \textsc{PersuasiveToM} benchmark along with open-source Llama-3 and Qwen2.5 models in strict compliance with their respective licenses and intended academic purposes. %We have manually inspected a subset of the data and found no personally identifying information or offensive content. 

% We conduct the human evaluation...
% Limitation of our work lies in the architectural variability of the intervention's impact. Our studies indicate that the effectiveness of \ourmethod interacts with the model's inherent representational patterns. The disparity in improvements magnitude suggests that the effectiveness of causal-oriented steering may interact with model-specific architecture priors. 
% Llama-3 appears higher sensitivity to our steering mechanism, whereas Qwen2.5's relative stability leads to more modest leaps. How diverse pre-training objectives and architectural designs influence a model's susceptibility to representational steering remains to be explored in future research. 
\section*{Acknowledgement}
This work was supported by the National Key Research and Development Program of China (Grant No. 2024YFB4505202), Major Program (JD) of Hubei Province (No. 2023BAA024), Singapore Ministry of Education (MOE)
Academic Research Fund (AcRF) Tier 1 grant (Proposal ID: 24-SIS-SMU-002) and China Scholarship Council.

\bibliography{latex/main} 

\appendix

\begin{figure*}
    \centering
    \includegraphics[width = 1.0\linewidth]{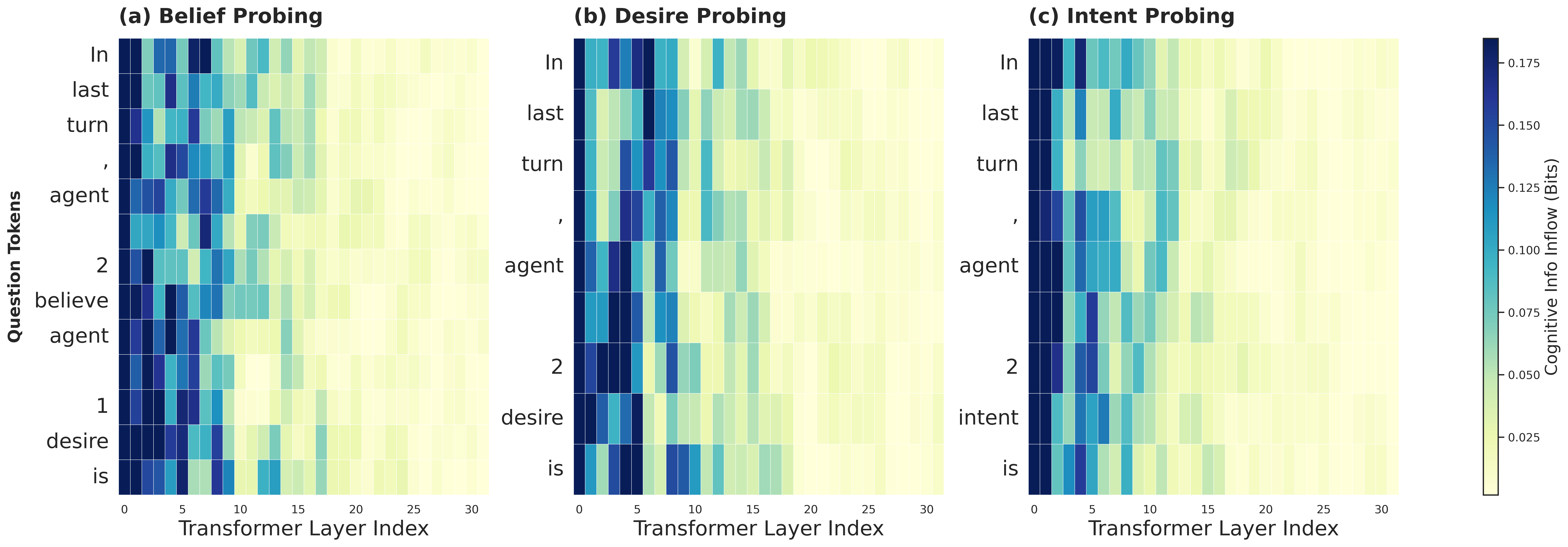}
    \caption{Layer-wise probing results on \textsc{NegotiationToM} with Llama3.}
    \label{heatmap-llama}
\end{figure*}

\begin{figure*}[h]
    \centering
    \includegraphics[width=1.0\linewidth]{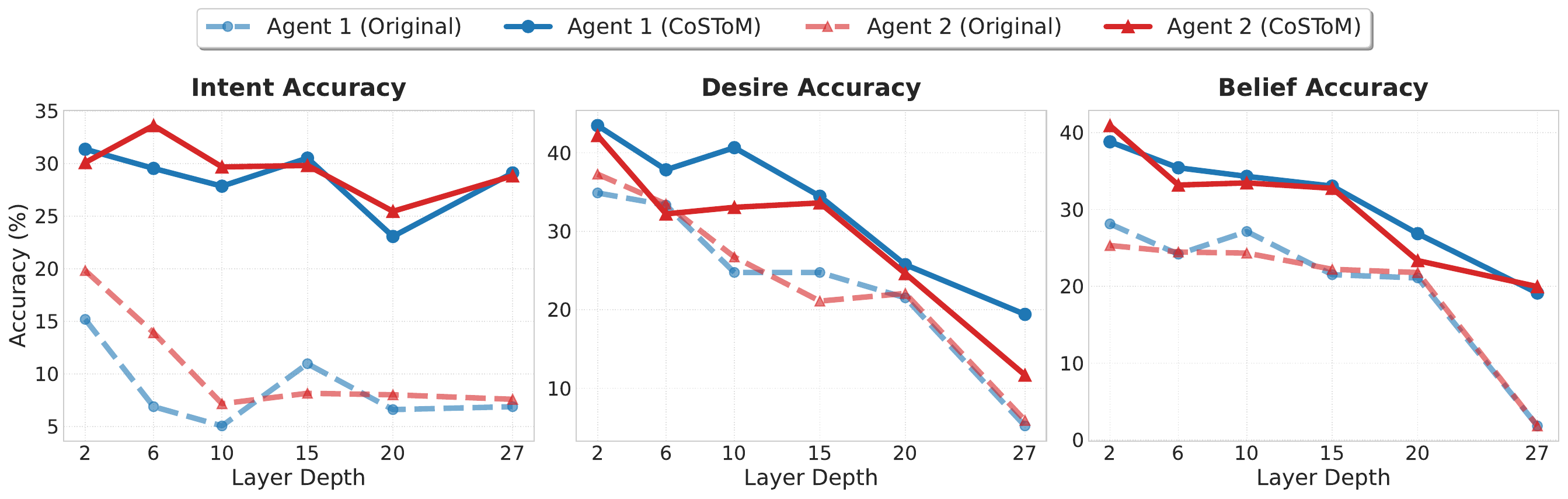}
    \caption{Layer-wise reconstruction accuracy on the \textsc{Negotiation} dataset (Qwen2.5).}
    \label{negotiation-intervene}
\end{figure*}
\begin{figure*}[h!]
    \centering
    \includegraphics[width=1.0\linewidth]{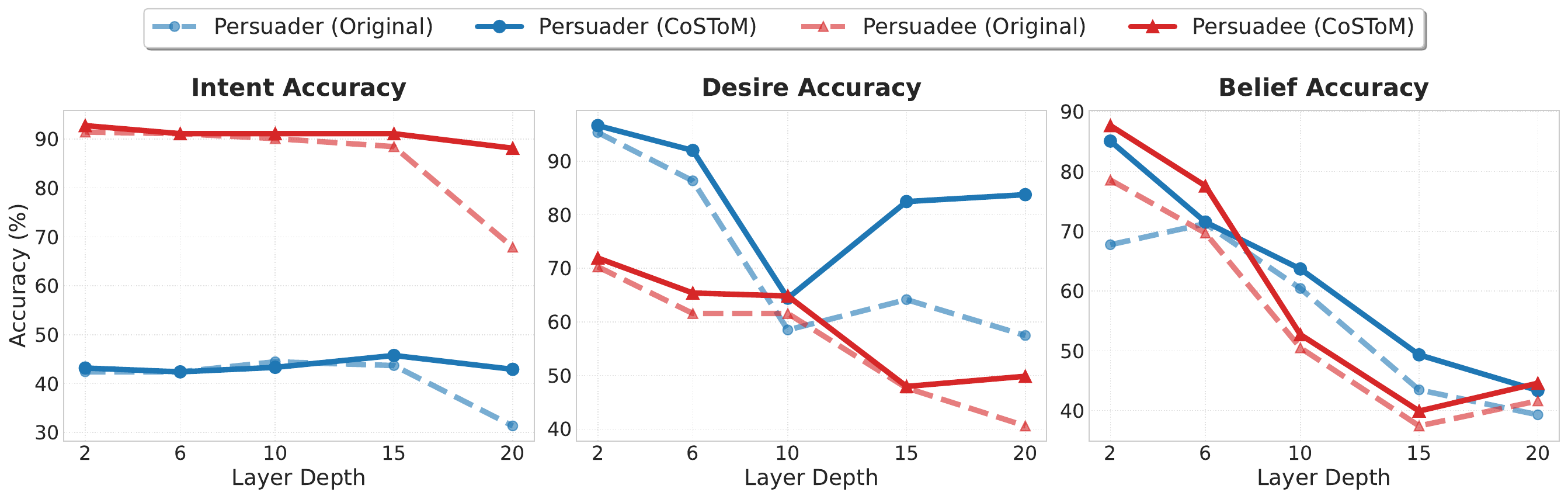}
    \caption{Layer-wise reconstruction accuracy on the \textsc{PersuasiveToM} dataset (Qwen2.5).}
\label{persuasive-intervene}
\end{figure*}

\begin{table*}[h]
\small
\centering
\begin{tabular}{l|ccc|ccc|ccc}
\toprule
\multirow{2}{*}{\textbf{Method}} & \multicolumn{3}{c|}{\textbf{Judge: Llama-3.3-70B}} & \multicolumn{3}{c|}{\textbf{Judge: GPT-5.1}} & \multicolumn{3}{c}{\textbf{Judge: Human}} \\
\cmidrule(lr){2-4} \cmidrule(lr){5-7} \cmidrule(lr){8-10}
 & \textbf{ToM} & \textbf{Coh.} & \textbf{PSE} & \textbf{ToM} & \textbf{Coh.} & \textbf{PSE} & \textbf{ToM} & \textbf{Coh.} & \textbf{PSE} \\
\midrule
\multicolumn{10}{c}{\textit{Base Model: Llama-3-8B-Instruct}} \\
\midrule
\textit{Prompting Baselines} \\
\hspace{3mm} Zero-shot Baseline &0.104 &0.433 &0.199 &0.259 &0.783  &0.529 &0.165 &0.520 &0.310\\

\hspace{3mm} MindDial (prompt) \cite{qiu2024minddial} &0.451 &0.554 &0.485 &0.374 &0.478 &0.377  &0.385 &0.535 & 0.420 \\
\midrule
\textit{Tuning \& Intervention} \\
\hspace{3mm} MindDial (Fine-tuned) \cite{qiu2024minddial} &0.643 &0.720 &0.645 &0.610 &0.747 &0.600 &0.590 &0.685 &0.615\\
\hspace{3mm} Full-Layer LoRA & 0.628 & 0.798 & 0.651 & 0.367 & \textbf{0.864} & 0.729 & 0.485 & 0.790 & 0.670\\
\hspace{3mm} LatentQA \cite{jafari2025beyond} &0.255 &0.324 &0.269 &0.300 &0.504 &0.298 &0.275 &0.440 &0.285 \\
\rowcolor{gray!10}
\hspace{3mm} \ourmethod (Ours) & \textbf{0.797} & \textbf{0.811} & \textbf{0.757} & \textbf{0.703} & 0.807 & \textbf{0.744} & \textbf{0.715} & \textbf{0.805} & \textbf{0.740} \\
\midrule

\multicolumn{10}{c}{\textit{Base Model: Qwen2.5-7B-Instruct}} \\ 
\midrule
\textit{Prompting Baselines} \\
\hspace{3mm} Zero-shot Baseline & 0.109 & 0.431 &0.190 &0.227 & 0.788 & 0.522 &0.145 &0.515 &0.305  \\

\hspace{3mm} MindDial (prompt)  \cite{qiu2024minddial} &0.442 &0.560 &0.455 &0.586 &0.667 &0.596 &0.495 &0.610 &0.515  \\
\midrule
\textit{Tuning \& Intervention} \\
\hspace{3mm} MindDial (Fine-tuned) \cite{qiu2024minddial}& 0.643&0.720 &0.645  &0.645 &0.723 &0.670 &0.635 &0.715 &0.655\\
\hspace{3mm} Full-Layer LoRA &0.654 &0.802 &0.636 &0.343  &0.817 &0.678 &0.445 &0.795 &0.660  \\
\hspace{3mm} LatentQA \cite{jafari2025beyond} &0.671 &0.751 &0.681 &0.630 &0.773 &0.638 &0.640 &0.765 &0.645\\
\rowcolor{gray!10}
\hspace{3mm} \ourmethod (Ours) & \textbf{0.802} & \textbf{0.811} & \textbf{0.713} & \textbf{0.746} & \textbf{0.888} & \textbf{0.812} & \textbf{0.760} & \textbf{0.840} & \textbf{0.795} \\
\bottomrule
\end{tabular}
\caption{Dialogue generation quality on the \textsc{PersuasiveToM} dataset (N = \(200\)). \ourmethod achieves the highest scores across nearly all metrics, validating its ability to enhance persuasion strategy effectiveness through accurate mental state attribution. (ToM: Theory of Mind Reasoning Quality, Coh.: Contextual Coherence, PSE: Persuasion Strategy Effectiveness.)}
\label{persuasion-generation}
\end{table*}
\begin{figure*}[t]
    \centering
    \begin{tcolorbox}[
        enhanced,
        sharp corners,
        boxrule=0.5pt,
        colback=gray!5,
        colframe=gray!80,
        fonttitle=\bfseries,
        title=LLM-as-a-Judge Scoring Rubric (Negotiation Task Example),
        fontupper=\small
    ]
You are a strict and critical evaluator in negotiation dialogues. You are provided with a Dialogue History and different model responses. Your task is to independently score EACH response against the criteria below (Scale: 0.0 to 1.0).

\vspace{0.5em}
\textbf{1. ToM Reasoning Quality}: Is the agent's understanding of the other's mental states accurate and appropriately explicit?
\begin{itemize}[leftmargin=1.5em, noitemsep, topsep=2pt]
    \item \textbf{1.0}: Highly accurate and explicit. Inference is fully grounded in the dialogue, and uses clear ToM language (e.g., "You believe that...").
    \item \textbf{0.8}: Mostly accurate and implicit. Core inference is sound, with minor over-interpretation, and uses implied ToM terms (e.g., "I understand...").
    \item \textbf{0.5}: Mixed accuracy. Half of the claims about the mental state are either unsupported or fabricated details.
    \item \textbf{0.2}: Major errors. Most inferred details are fabricated (e.g., "your son has asthma" when not mentioned).
    \item \textbf{0.0:} Completely fabricated mental states or no psychological phrasing used at all.
\end{itemize}

\textbf{2. Contextual Coherence}: Is the response logically and topically aligned with the dialogue history, and are proposals/reasons grounded in the known facts?
\begin{itemize}[leftmargin=1.5em, noitemsep, topsep=2pt]
    \item \textbf{1.0}: Fully coherent and grounded. Response is a logical continuation, and all proposals/reasons are directly supported by the dialogue history.
    \item \textbf{0.8}: Well-aligned. Response is logically sound but may contain minor, non-critical conversational redundancy or external details. 
    \item \textbf{0.5}: Partially disconnected. Response addresses the immediate previous turn but introduces a new, irrelevant topic or resource that lacks clear context.
    \item \textbf{0.2}: Logical error. Proposal contradicts established facts or resources known from the dialogue (e.g., trading for a resource known to be near a stream).
    \item \textbf{0.0}: Totally disjointed. Response repeats history or fails to address the previous turn.
\end{itemize}

\textbf{3. Negotiation Strategy Effectiveness}: Does the response constructively advance the deal by offering balanced proposals, logical counter-arguments, or maintaining a cooperative frame?
\begin{itemize}[leftmargin=1.5em, noitemsep, topsep=2pt]
    \item \textbf{1.0}: Highly effective. Proposes a new, **concrete, and balanced trade-off solution**, framed using highly cooperative language.
    \item \textbf{0.8}: Constructive response. Clearly accepts/refutes the previous offer with a logical justification, maintaining a high to medium cooperative tone.
    \item \textbf{0.5}: Passive response. Merely confirms the previous statement or expresses vague wishes ("sounds fair"), without actively moving the negotiation forward.
    \item \textbf{0.2}: Zero-sum/Stalling. Focuses only on self-interest, refuses reasonable compromise, or attempts to stall the negotiation.
    \item \textbf{0.0}: Negotiation breakdown. Uses antagonistic language or proposes obviously unacceptable terms.
\end{itemize}
    \end{tcolorbox}
    \caption{LLM-as-a-Judge Scoring Rubric for Negotiation Task.}
    \label{rubric negotiation}
\end{figure*}

\begin{figure*}[!t]
    \centering
    \begin{tcolorbox}[
        enhanced,
        sharp corners,
        boxrule=0.5pt,
        colback=gray!5,
        colframe=gray!80,
        fonttitle=\bfseries,
        title=LLM-as-a-Judge Scoring Rubric (Persuasive Task Example),
        fontupper=\small
    ]
You are a strict and critical evaluator in persuasive dialogues. You are provided with a Dialogue History and different model responses. Your task is to independently score EACH response against the criteria below (Scale: 0.0 to 1.0).

\vspace{0.5em}
\textbf{1. ToM Reasoning Quality}: Accuracy of mental state inference
\begin{itemize}[leftmargin=1.5em, noitemsep, topsep=2pt]
    \item \textbf{1.0}:  Perfectly infers desire/belief/intent from dialogue; uses explicit ToM language ("you believe...", "your concern is...")
    \item \textbf{0.8}: Accurate inference, implicit phrasing ("I see you value..."), demonstrates social awareness without formalizing it
    \item \textbf{0.5}: the response identifies some mental states correctly but includes 1-2 unsupported or speculative details (social hallucination)
    \item \textbf{0.2}: incorrectly assigns preferences or intentions that contradict the dialogue history
    \item \textbf{0.0:} provides a generic response that ignores the partner's psychological state entirely
\end{itemize}

\textbf{2. Contextual Coherence}: Discourse flow, factual grounding, and relevance.
\begin{itemize}[leftmargin=1.5em, noitemsep, topsep=2pt]
    \item \textbf{1.0}: Seamless discourse integration, response is perfectly grounded in prior facts with natural flow and zero logical redundancy
    \item \textbf{0.8}: Strong alignment, logically sound but contains minor repetitive phrasing or slight conversational fluff
    \item \textbf{0.5}: Surface-level coherence, follows basic turn-taking rules but feels formulaic (e.g., ``I understand you feel X, let's do Y'') and lacks deep topical nuance
    \item \textbf{0.2}: Factual inconsistency, contradicts established history or introduces resources/facts not present in the context
    \item \textbf{0.0}: Discursive breakdown, incoherent, off-task, or fails to respond to the immediate previous turn
\end{itemize}

\textbf{3. Persuasion Strategy Effectiveness}: move persuasion forward
\begin{itemize}[leftmargin=1.5em, noitemsep, topsep=2pt]
    \item \textbf{1.0}: Proposes a highly compelling argument tailored to the partner's specific concerns. Uses advanced techniques (e.g., "foot-in-the-door," emotional storytelling, or expert social proof) with high empathy
    \item \textbf{0.8}: Provides logical justifications or clear emotional appeals. Directly addresses the partner's stance and offers a solid reason to reconsider, maintaining a respectful and encouraging tone
    \item \textbf{0.5}: Uses canned persuasive slogans or vague moralizing ("It's for a good cause") without addressing the specific dialogue context. Passive and unlikely to change a firm stance
    \item \textbf{0.2}: Dismisses the partner's objections or uses a condescending tone ("You are wrong to think that"). Likely to trigger psychological reactance (making the partner more stubborn)
    \item \textbf{0.0}: Hostile language, aggressive pressure, or a complete failure to address the persuasive goal
\end{itemize}
    \end{tcolorbox}
    \caption{LLM-as-a-Judge Scoring Rubric for Persuasive Task.}
    \label{rubric persuasive}
\end{figure*}

\section{Implementation Details}
\label{implementation details}
\subsection{Hardware and Software Environment}
Our experiments were conducted using Llama-3-8B-Instruct (approximately \(8.03\) billion parameters) and Qwen2.5-7B-Instruct (approximately \(7.61\) billion parameters) as base models. The computational framework was implemented using PyTorch 2.9.1 and the HuggingFace Transformers/PEFT libraries. All experiments were conducted on a high-performance computing node equipped with four NVIDIA L40S GPUs (\(48\)GB of GDDR6 VRAM each). To manage the memory footprint of the dual-model architecture, we leveraged the QLoRA framework \cite{dettmers2023qlora}, employing NormalFloat 4(NF4) as the storage data type and BFloat16 (BF16) as the compute data type to maintain numerical stability during the gradient-bridge backpropagation.
% To manage the memory footprint of the dual-model architecture, we utilize BFloat16 (BF16) mixed-precision training across all stages.
\subsection{Training and Hyperparameters}
For the causal-oriented steering, we applied LoRA (Low-Rank Adaptation) specifically to the ToM-critical layers identified in Section \ref{reading}. We targeted all linear modules within the Transformer blocks, including the attention projections ($q\_proj, k\_proj, v\_proj, o\_proj$) and the feed forward network layers (\(gate\_proj, up\_proj, down\_{proj}\)). To ensure reproducibility, we fixed the random seed to \(42\) for all initialization and data sampling. 
The specific hyperparameters used for training are summarized as follows:
\begin{itemize}[leftmargin=1em]
    \item \textit{Optimization}: we employed the \textbf{AdamW} optimizer with a linear learning rate scheduler and a peak learning rata of $1\mathrm{e}{-4}$.
    \item \textit{LoRA Settings}: The LoRA rank \(r\) was set to \(16\) with an alpha parameter \(\alpha = 32\). We applied a LoRA dropout of \(0.05\) to mitigate overfitting.
    \item \textit{Training Dynamics}: Training was conducted with a batch size of \(4\) per GPU. While the maximum number of epochs was set to \(10\), we implemented an \textbf{early stopping mechanism} with a patience of \(3\) epochs. 
    \item \textit{Convergence}: Early stopping was triggered if the validation loss failed to improve by more than \(0.01\) (threshold), or if the absolute loss fell below a \textbf{minimum threshold} of \(0.1\).
\end{itemize}

\subsection{Evaluation Settings}
For the LLM-as-a-Judge evaluation, we employed GPT-5.1 and Llama-3.3-70B-Instruct via OpenAI API with a temperature of \(0.0\) to minimize variance in scoring. All prompts used for generation and evaluation are detailed in Appendix \ref{rubrics}.

\section{ToM Interpretation Analysis (RQ1)}
\label{persuasive-interpretation}
% Table \ref{persuasive_tracing} presents the detailed causal tracing results of the \textsc{PersuasiveToM} dataset. 
Figure \ref{heatmap-llama} demonstrates the layer-wise results on the \textsc{NegotiationToM} dataset with Llama-3 model. 
We conduct layer-wise robing by training linear classification on hidden representation to predict ToM categories. Following \cite{DBLP:conf/coling/Ju0DYRL24}, we use $\mathcal{V}$-usable information rather than raw accuracy to measure knowledge decodability. High $\mathcal{V}$-usable values indicate a high concentration of accessible ToM-specific knowledge at a particular layer.

\section{Effectiveness of \ourmethod (RQ2)}
\label{qwen-intervention} 
To demonstrate the architectural robustness of \ourmethod, Figure \ref{negotiation-intervene} and \ref{persuasive-intervene} illustrate the impact of causal-oriented steering on the Qwen2.5 model. These visualizations confirm the effectiveness of \ourmethod in mitigating representation collapse and enhancing the BDI decodability generalizes across different LLM families.

\section{Comparative Analysis of Dialogue Generation (RQ3)}
\label{generation-compare}
We provide a comprehensive comparison between \ourmethod and five baselines methods regarding dialogue generation quality on the \textsc{PersuasiveToM} dataset. Detailed performance results are documented in Table \ref{persuasion-generation}. 

\section{Task-specific Instruction Prompt}
\label{instruction input}
We employ different prompting strategies tailored to the architectural requirements of the evaluation methods.

\noindent\textbf{Baseline Paradigms (Zero-shot and Full-Layer LoRA)}: For these single-model baselines, we concatenate the \textit{dialogue history} and the \textit{instruction prompt} into a single input sequence.

\noindent\textbf{Dual-model Framework (\ourmethod and LatentQA)}: In our dual-model setting, we decouple the inputs: the \textit{dialogue history} is fed into the context encoder for mental state representation, while the task-specific \textit{instruction prompt} is provided to the decoder to guide the response generation. Thus, as a plug-and-play module, \ourmethod-enhanced model can be adapted into diverse downstream interaction tasks. 

\noindent\textbf{Two-stage Pipeline (MindDial)}: MindDial \cite{qiu2024minddial} follows an \textit{inference-then-generation} pipeline. In stage 1, the model reason over the BDI states based on the dialogue history to produce \textit{ToM analysis results}. In stage 2, these results are integrated into the system prompt to generate the final response. 

\vspace{1em}
\noindent\textit{Instruction Prompt Example (Negotiation Task)}:
\begin{quote}
\small\texttt{
You are an agent in a cooperative negotiation about trip resources (Food, Firewood, Water). Based on the conversation history, give the next utterance, considering the other agent's desires, believes, and intends, even those are not explicitly stated. Respond in a way that shows you understand their perspective and reaches a agreement.
}  
\end{quote}

\noindent\textit{Instruction Prompt Example (Persuasive Task)}:
\begin{quote}
\small\texttt{
You are the persuader in a two-person dialogue. Your goal is to generate the next response in the conversation to successfully persuade the other person (the persuadee). Based on the conversation history, infer the persuadee's desires, believes, and intends, craft a single, continuous persuasive response.
}  
\end{quote}

\section{Dialogue Generation Evaluation Rubrics}
\label{rubrics}
To ensure a rigorous and reproducible assessment of the generated dialogues, we utilize expert-designed rubrics across three functional axes, and each response is independently adjudicated on a \(0.0\) to \(1.0\) scale by the automated LLM evaluator. Below we provide the detailed prompt template and scoring criteria. 
% (i) ToM-centric metrics: \textit{ToM Reasoning Quality}, (ii) dialogue-level metrics: \textit{Contextual coherence}, and (iii) Objective-oriented metrics: \textit{Strategy Effectiveness}. 
\subsection{Details of Human Evaluation}
To ensure the high quality of the evaluation, we recruited \(3\) human annotators. All participants are graduate students in NLP with a strong understanding of conversational systems and Theory-of-Mind concepts. All annotators are proficient in English and were recruited from our internal university network.

Before the formal evaluation, we conducted a 30-minute training session to familiarize them with the scoring rubrics and provide anchor examples for each score level. The specific scoring rubrics are identical to those used for the LLM judge as shown in section \ref{llm-judge}.

For each of the \textsc{NegotiationToM} and \textsc{PersuasiveToM} test sets, we randomly sample \(100\) and \(200\) dialogues, respectively, stratified by interaction stage (\textit{beginning} : \textit{middle} : \textit{final} = \(1:2:1\)). For each instance, annotators were presented with: 

\textbf{Dialogue History}: The full context of the interaction.

\textbf{Model Responses}: Anonymized and randomized responses generated by \ourmethod and baselines. The order of the models was shuffled for each instance to eliminate position bias.

To validate the reliability of the human scores, we calculated the inter-annotator agreement using Fleiss' Kappa \cite{fleiss1971measuring}. The average Kappa scores across the three metrics were:
\textbf{ToM Reasoning Quality}: $\kappa = 0.72$, \textbf{Contextual Coherence}: $\kappa = 0.81$, \textbf{Strategy Effectiveness}: $\kappa = 0.68$. These values indicate a substantial level of agreement among the annotators, confirming the robustness of our human evaluation results.
 
\subsection{LLM-as-a-Judge Instruction}
\label{llm-judge}
Figure \ref{rubric negotiation} and \ref{rubric persuasive} present the detailed evaluation rubrics and prompting instructions for the LLM-as-a-Judge framework on the negotiation and persuasion tasks, respectively.

\end{document}